# Entropy of Telugu

Venkata Ravinder Paruchuri
Computer Science
Oklahoma State University, Stillwater, OK-74078
venkarp@okstate.edu

**Abstract**: This paper presents an investigation of the entropy of the Telugu script. Since this script is syllabic, and not alphabetic, the computation of entropy is somewhat complicated.

## Introduction

Indian languages are highly phonetic; i.e. the pronunciation of new words can be reliably predicted from their written form. A background to Indian scripts is given in [1]-[6]. Indian scripts are highly systematic in their arrangement of sounds, and the milieu in which they arose is provided in papers on early Indian science [7]-[11]. From the evidence available at this time it may be assumed that the 3$^{rd}$ millennium BC Indus script evolved into the Brahmi script of late centuries BC which, in turn, evolved into the different Indian modern Scripts. Structurally, all the Indian scripts are thus closely related although their forms may look quite different. The Brahmi script is also the parent to Southeast Asian scripts.

The alphabets of Indian languages are classified into consonants, vowels and other symbols. An akshara or syllable consists of 0, 1, 2 or 3 consonants and a vowel or other symbol. Each akshara can be pronounced independently. Alphabets of all the Indian languages are derived from the Brahmi script. In all the Indian languages there are 33 common consonants and 15 common vowels. In addition, there are 3-4 consonants and 2-3 vowels that are specific to each language, but not very significant in practice. Words are made up of one or more aksharas. If an akshara consists of more than one consonant, they are called samyuktakshara.

The similarity in alphabet is not extended in graphical form which is used for printing. Each language uses different scripts, which consists of different graphemes. There are about 10-12 major scripts in India among which Devanagari is the most widely used. Different languages have different statistical characteristics. Some have an over line for the entire word, while some have not touching graphemes. The vowels and the supporting consonants in samyuktakshara can appear on the left, right, above, below or in combinations to the main consonant.

Examples:

| | |
|---|---|
| Over line | जम्मू-कश्मीर |
| Non Touching | పద్మవిభూషణ్ |

## Entropy

Entropy of the language is the measure of disorder associated with the system. For a random variable X with n outcomes $\{x_i : i = 1,2,3 \ldots n\}$, the Shannon entropy, a measure of uncertainty and denoted by H(X), is defined as

$$H(X) = -\sum_{i=1}^{n} x_i \log_b x_i$$

Entropy of the English language is calculated by taking into consideration, the 26 alphabets and space character and leaving out the punctuation.

Entropy of Telugu language is computed by converting the Telugu language into English language using some software's and then using the above mentioned formula to compute it. The entropy in Telugu language is computed in two ways.

- By converting it to English and then considering them as English letters
- By converting it to English and then considering them as Telugu letters.

To understand the conversion of Telugu into English font, here are some of the examples:

| | |
|---|---|
| కార్యాలయం | kAryAlayaM |
| ప్రతిపాదించినట్లు | pratipAdiMcinaTlu |
| పద్మవిభూషణ్ | padmavibhUShaN^ |
| వయస్సు | vayassu |
| సాధ్యమైనంత | sAdhyamainaMta |

In the first method, after converting them into English font, they are considered as English alphabets, i.e. padmavibhUShaN^ is considered as a sequence of characters containing p, a, d, m, a, v, i, b, h, U, S, h, a, N, ^. In the Telugu language, the alphabets are case sensitive as each has different meaning. The frequencies of each letter are shown in the table 1.

Table 1: Frequencies of alphabets in Telugu language

| Alphabets | Frequencies (%) | Alphabets | Frequencies (%) |
|---|---|---|---|
| A | 13.15 | A | 6.13 |
| B | 0.99 | B | 0.00 |
| C | 1.98 | C | 0.00 |
| D | 2.47 | D | 1.35 |
| E | 1.30 | E | 1.77 |
| F | 0.04 | F | 0.00 |
| G | 1.52 | G | 0.00 |
| H | 3.31 | H | 0.00 |
| I | 6.81 | I | 1.03 |
| J | 0.78 | J | 0.04 |
| K | 3.92 | K | 0.00 |
| L | 3.57 | L | 0.69 |
| M | 2.01 | M | 3.67 |
| N | 4.66 | N | 0.25 |
| O | 0.49 | O | 1.46 |
| P | 2.68 | P | 0.04 |
| Q | 0.00 | Q | 0.01 |
| R | 3.31 | R | 0.06 |
| S | 2.46 | S | 0.29 |
| T | 2.97 | T | 1.41 |
| U | 5.65 | U | 0.07 |

| V | 2.43 | V | 0.01 |
|---|------|---|------|
| W | 0.00 | W | 0.00 |
| X | 0.00 | X | 0.00 |
| Y | 1.86 | Y | 0.00 |
| Z | 0.00 | Z | 0.00 |
| Space | 12.05 | ^ | 0.46 |

The above results are calculated based on 10,000 characters and the frequencies are rounded off to 2 decimal points.

In the next method, after converting them to English font, they are considered as Telugu syllables as opposed to English alphabets in method one. In this approach, the words are partitioned on the basis of Telugu syllables. Some examples are shown below.

        padmavibhUShaN^        pa, dma, vi, bhU, Sha , N^

        kAryAlayaM        kA, ryA, la, yaM

        sAdhyamainaMta        sA, dhya, mai, naM, ta

The frequencies of each syllable according to the given text are computed and then the entropy of the language is calculated.

In this case we are considering one syllable at a time and the entropy of the language calculated is approximately 5.98 for the 10,000 characters that I have considered.

We have continued this method for finding the entropy of Telugu language by considering two syllables at a time to decrease the entropy of the language.

In this approach, the words are partitioned on the basis of Telugu syllables. Some examples are shown below.

        padmavibhUShaN^        padma, dmavi, vibhU, bhUSha, ShaN^

        kAryAlayaM        kAryA, ryAla, layaM

        sAdhyamainaMta        sAdhya, dhyamai, mainaM, naMta

In this case the entropy of the language is calculated to be 3.98 approximately for the same 10,000 characters.

In the next step, we found the entropy of Telugu language by considering three syllables at a time. In this approach, the words are partitioned on the basis of Telugu syllables. Some examples are shown below.

|  |  |
|---|---|
| padmavibhUShaN^ | padmavi, dmavibhU, vibhUSha, bhUShaN^ |
| kAryAlayaM | kAryAla, ryAlayaM |
| sAdhyamainaMta | sAdhyamai, dhyamainaM, mainaMta |

In this case the entropy of the language is calculated to be 2.739 approximately for the same 10,000 characters.

In the next step, we found the entropy of Telugu language by considering four syllables at a time. In this approach, the words are partitioned on the basis of Telugu syllables. Some examples are shown below.

|  |  |
|---|---|
| padmavibhUShaN^ | padmavibhU, dmavibhUSha, vibhUShaN^ |
| kAryAlayaM | kAryAlayaM |
| sAdhyamainaMta | sAdhyamainaM, dhyamainaMta |

In this case the entropy of the language is calculated to be 2.077 approximately for the same 10,000 characters.

We continued this process for up to six syllables and the entropy of the language when 5 syllables and 6 syllables are considered is calculated.

The entropy of language for 5 syllables is 1.699. The entropy of language for 6 syllables is 1.39. All the results are represented diagrammatically using graphs in the results section.

## Results

This graph compares the entropies of English and Telugu language. Entropy of the English language is calculated by taking into consideration, the 26 letters and space character and leaving out the punctuation.

The entropy in Telugu language is computed in two ways.

- By converting it to English and then considering them as English alphabets (Telugu 1).
- By converting it to English and then considering them as Telugu alphabets (Telugu 2).

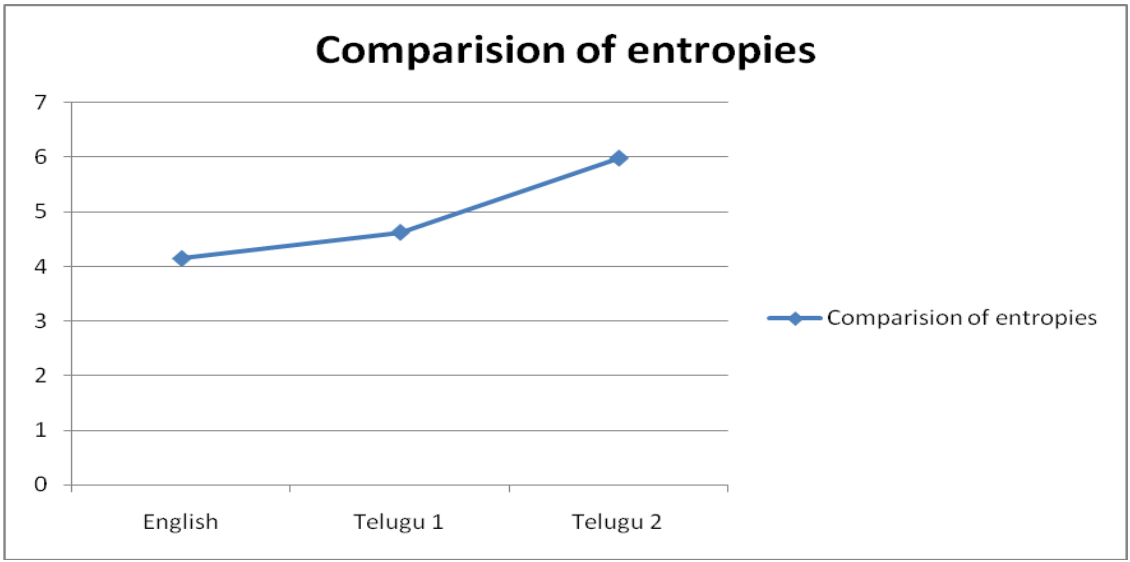

Graph 1: Comparison of entropies calculated.

The next section is about Telugu 2 where the alphabets are considered as in Telugu language. We have considered one syllable at a time, two syllables at a time etc till we have reached a value which doesn't change much. Entropy decreases as the number of syllables considered increases.

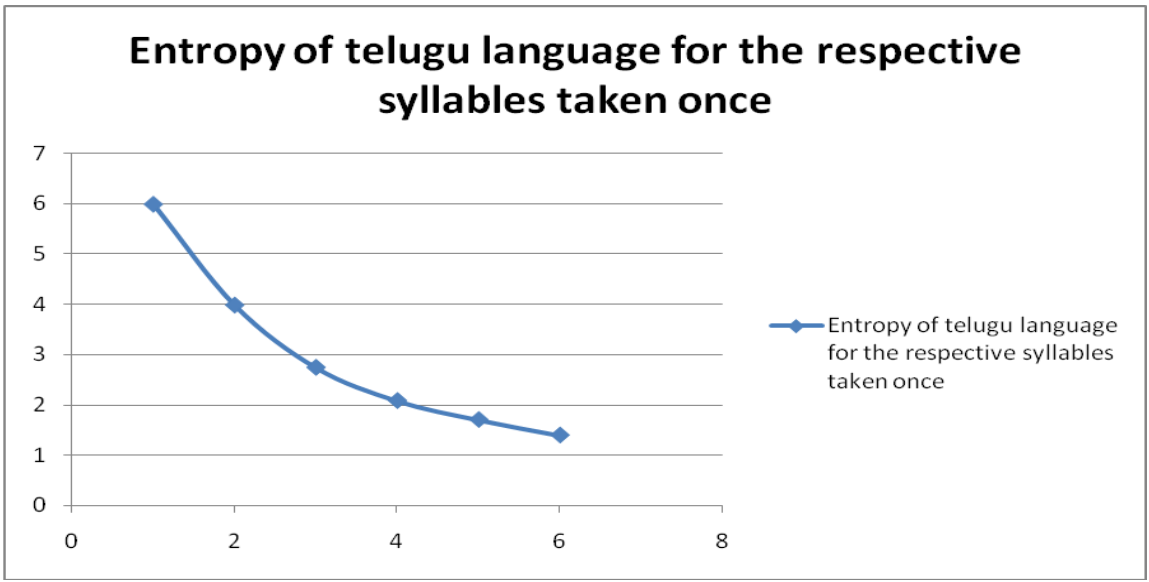

Graph 2: Entropy of language when considered according to syllables.

The next graph shows the comparison of entropies of English language for the original text and the jumbled text [12] with first and last letters kept constant. This graph shows that there is much less difference in the entropies of original text and the jumbled one and hence people are able to read the jumbled text with not much difficulty.

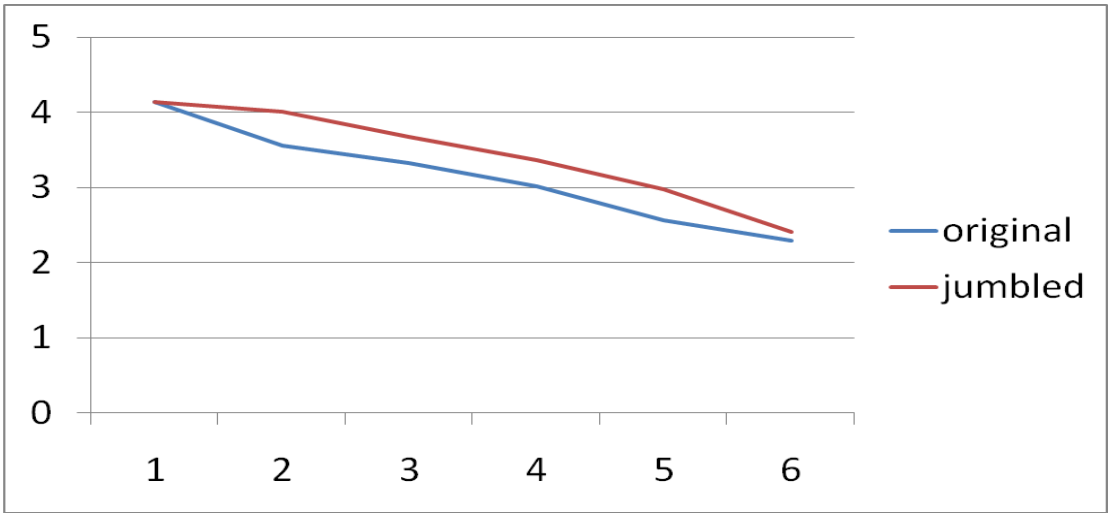

Graph 3: Entropies of English language for original text and jumbled text.

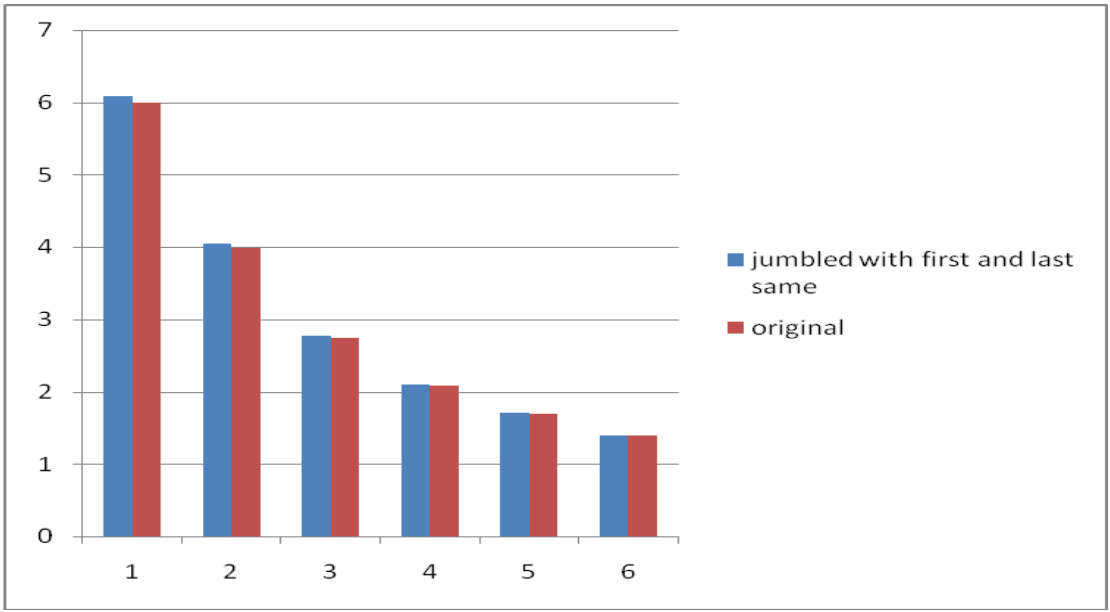

Graph 4: Entropies of English language for original text and jumbled text.

The above graph shows the comparison of entropies of Telugu language for the original text and the jumbled text with first and last letters kept constant. This graph shows that there is much less difference in the entropies of original text and the jumbled one and hence people are able to read the jumbled text with not much difficulty.

The difference is very little in Telugu language compared to that of English is because of the word length in Telugu language. In jumbling process we only consider the words of length greater than three, but the graph shows that approximately 70% of the words in Telugu are of length less than four.

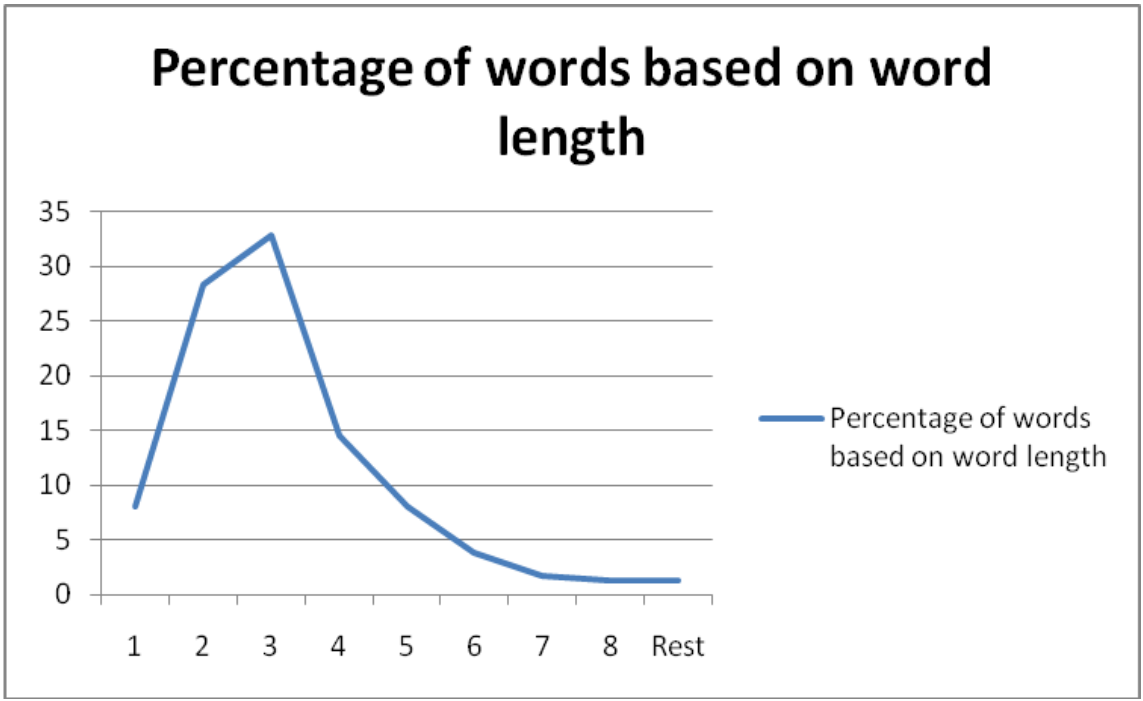

Graph 22: Percentage of words based on word length

## Conclusions

For Indian languages like Hindi and Telugu, which are phonetic so that the pronunciation of new words can be reliably predicted from their written form, consonants carry more information than the vowels. This can be inferred from both the time taken to read the text and also from the percentage of correctness. The words can be read with much ease if the first and last letters are left in their places.

The entropy of Telugu is higher than that of English, which means that Telugu is more succinct than English and each syllable in Telugu (as in other Indian languages) contains more information compared to English. The comparison of entropies of the language for original and jumbled texts confirms that the users can read jumbled text with not much difficulty.